\documentstyle[colap]{article}

\title{\vspace{-0.5in}Using sentence connectors for evaluating MT output}
\author{Eric M. Visser$^{0}$ and Masaru Fuji\vspace{3mm}\\
Fujitsu Laboratories Ltd., Media Integration Laboratory\\4-1-1
Kamikodanaka, Nakahara-ku, Kawasaki 211,
Japan\\{\{}eric$\mid$fuji{\}}@flab.fujitsu.co.jp}

\begin{document}

\bibliographystyle{acl}

\maketitle
\vspace{-0.5in}
\begin{abstract}

This paper elaborates on the design of a machine translation
evaluation method that aims to determine to what degree the meaning
of an original text is preserved in translation,
without looking into the grammatical correctness of its constituent
sentences.
The basic idea is to have a human evaluator take the sentences of the
translated text and,
for each of these sentences,
determine the semantic relationship that exists between it and the
sentence immediately preceding it.
In order to minimise evaluator dependence, relations between sentences
are expressed in terms of the {\bf conjuncts} that can connect them,
rather than through explicit categories.
For an $n$-sentence text
this results in a list of $n-1$ sentence-to-sentence relationships,
which we call the text's {\bf connectivity profile}.
This can then be compared to the connectivity profile of the original
text, and the degree of correspondence between the two would be a
measure for the quality of the translation.

A set of ``essential'' conjuncts was extracted for English and
Japanese, and a computer interface was designed to support the task of 
inserting the most fitting conjuncts between sentence pairs.
With these in place, several sets of experiments were performed.

\end{abstract}

\section{Background}

\footnotetext{The first author is currently at ATR Interpreting
Telecommunications Research Laboratories; current e-mail address is
(eric@itl.atr.co.jp).}
Evaluation of MT results is generally tackled on a very detailed,
linguistic-technical level.
Typically, a test set of sentences is prepared each of which is
carefully designed to ascertain whether the MT system can handle a
certain grammatical phenomenon  --- e.g.\ \cite{Isahara:MTS95}.
Other methods may concentrate on word choice,
consistency in terminology, PP attachment, dependency relations,
or other specific grammatical or lexical aspects.
While such evaluation methods are certainly necessary and useful for
the MT developer, they do not necessarily give us a reliable
indication of user satisfaction.

Especially now that MT systems are becoming widely available on the
home user market and coming within the casual user's reach,
MT developers need to pay more attention to this aspect.
Casual users might just not care all that much about grammatical
correctness: as long as they can {\em understand} the output, they
might be satisfied with the system.
Moreover, such users are not likely to judge the system on a
sentence-by-sentence basis: rather, they will be interested in the
understandability of the text as a whole.
The flurry of integrated WWW-browsers cum MT systems\footnote{These
allow you to read English WWW-pages in Japanese, preserving the
original page layout.}
to hit the (Japanese) market recently has added to the plausibility of
this scenario.

We conclude then that an MT evaluation method is called for which
concentrates on whole texts rather than on single sentences,
and which judges meaning and readability rather than grammar.
In addition, we specify that evaluation results should be
reproducible and evaluator-independent (to a reasonable degree
at least), and quantifiable.
These additional requirements are necessary to ensure that results
obtained at different times and/or by different evaluators (preferably
using different texts) are comparable.

In \cite{Su:COLING92} an interesting alternative evaluation method is
proposed, in which the discrepancy is measured between raw MT output
and the post-edited result.
This method does work on whole texts,
and could conceivably be adapted to judge meaning and readability
(by adequately instructing the post-editors);
then again in ``browsing'' applications post-editing is not the norm,
and it may be difficult to attain a good approximation of
``browsable'' MT output.
In this paper, we try a different approach.

\section{Outline of the evaluation method}

\subsection{Compare salient properties}

To test whether the meaning of a translated text has come across, one
could simply ask the evaluators questions about the translated text,
or have them summarise it.
Such methods however are either costly (for each new text a new set of 
questions will have to be devised) or hard to quantify objectively, or
even both.

The method we will adopt involves constructing a profile of both the
original and the translated text in terms of some salient semantic or
pragmatic property of its constituent sentences.
These profiles can then be compared to give an indication of
translation quality:
if we assume that the original text's profile is ``perfect'',
then the degree to which the profile of the translated text resembles 
the perfect profile will correspond
(in theory at least) to the quality of the translation.
This approach assumes that the number and order of sentences are
invariant in translation;
luckily, for MT systems, this is almost always true.

As for the salient property to be used in the profile,
we settled on
meaning relations of single sentences with previous text:
this property seemed to us to be both fairly discriminating
and implementable.
In summary, a profile will be an ordered list of
meaning relations $x_{i}$, i = 2 \ldots\ n
which describe the relation of sentence i with what came before.
Moreover, the target of each relation is taken to be
the previous sentence, i.e.\ sentence i-1 (see \S~\ref{Imp} for
further discussion).

\subsection{Avoid contrived definitions}

A set of sentence-to-sentence relation categories  will then have to
be designed and defined; but the wide variety of proposed methods and
solutions (see \cite{Hovy:1993} for an overview) suggests that this is
not an easy task.
Indeed, the problem with categories and definitions is that
the evaluator will always have to depend to a certain extent on
his own personal understanding of these definitions;
and the more categories there are,
the greater the chance that their definitions will not always be clear
and fixed in his mind.
This naturally has a deleterious effect on the reliability and
universality of evaluation results.

We will get back to the design problem later,
but with respect to the definition problem,
our solution was to simply hide the definitions.
We have sought to accomplish this by instructing the evaluator to
link sentences linguistically; more specifically, we have opted to
instruct the evaluator to choose
a {\bf conjunct}\footnote{A subclass of the adverbs,
cf.\ \cite{Quirk:GEngl} pg.\ 631-.
For languages that do not recognise this class, surrogates can be 
concocted: for Japanese, a mixture of conjunctions and conjoining adverbs.}
to be inserted between every pair of consecutive sentences.
The conjuncts themselves may be divided into categories,
but these can remain hidden from the evaluator.
This approach hinges on the hope that straight linguistic
knowledge comes more naturally to people and is less susceptible to
person-to-person differences than contrived meaning categories.

\subsection{Standardise thinking methods}

Small-scale preliminary experiments (on paper) showed that in spite of 
the above refinements, evaluator differences were still larger
than seemed reasonable.
We surmised that this was due to differences in work methods (or
thinking methods),
and that therefore these needed to be equalised a little more.
We decided on two countermeasures.

Recognising that the class of conjuncts was too large for the
evaluator to encompass at a glance,
we decided to implement an interactive Q{\&}A interface on the
computer in order to gradually guide the
evaluator to the optimal choice of a conjunct.
Obviously this opens a whole new can of worms, in that the interface
has to be designed (the kind and order of questions etc.); we will get
back to that later (in \S~\ref{Imp}).

The other step was to instruct the evaluator to extract the {\bf
topic} and {\bf comment} of the sentence under consideration.
Both topic and comment were only loosely defined: in truth the topic
and comment are not important as such, rather their extraction was
intended as a means to force the evaluator to get a clearer picture of
the meaning of the sentence under consideration
(though we did not tell them this).

\section{Basic assumptions}

At this point, it is useful to look back at the design considerations
outlined above and to clarify exactly what assumptions on sentences
and relations underlie them.
With a little luck, our results can provide some support for these
assumptions.

The first of our assumptions is that it is always possible to make
explicit the relationship of a sentence to what has come before using
a conjunct.
The conjunct may be present in the sentence, but even if it is not, it
can be added in a linguistically satisfactory way.
We also assume that the assignment of acceptable conjuncts is
reader-independent to a large degree.

We assume that conjuncts (which form a closed class) can be
divided into a limited number of categories that are meaningful
in terms of expressing the semantic relationship between sentences.

Yet another assumption is that the meaning relationships between
sentences of a text combine to form a characteristic feature (a
profile) of that text, and that this profile needs to be preserved in
translation.
Moreover, the ease with which this profile can be
discerned in the translated text is assumed to be related to the
readability or understandability of the text as a whole.

\section{The implementation\label{Imp}}

A prototype was implemented on a Macintosh computer using HyperCard.
The evaluation process is made up of the following steps,
which have to be executed for every sentence in the text.

\begin{enumerate}
\item Extract the topic(s) and comment(s) of the sentence under
consideration.
\item If there is more than one topic/comment pair, order the
pairs as seems best and determine (using the same method as for
sentences) which conjuncts fit best between the pairs.
\item Determine through a dialog with the system
which conjunct fits best at the start of the sentence under
consideration.
\end{enumerate}

\noindent
A backtrack function was implemented which allowed the subjects
to come back on decisions made earlier in the dialog.
The prototype keeps a very detailed log of what the evaluator does
exactly.
Without going into technical details, the following were the main
tasks in the implementation.

\mbox{}\\{\bf Categorising the conjuncts}\\
Our first categorisation of conjuncts was based on information
concerning conjuncts and rhetorical structures that we patched
together from authoritative grammars
for English \cite{Quirk:GEngl}, Japanese \cite{Martin:GJpn} e.a.
We came up with 9 categories; in a later redesign
we took the conjuncts themselves as our starting point and, by tracing
crossreferences in dictionaries, were able to reduce the initial
number of $\pm$ 220 to 32 ``basic'' conjuncts, divided over 11 
categories.

\mbox{}\\{\bf Assisting topic/comment extraction}\\
Frankly we have been unable to find a foolproof method, and have
settled for user-requested online help cued on linguistic aspects of
the sentence.

\mbox{}\\{\bf Defining the scope of meaning relations}\\
We have established above that meaning relations hold between
consecutive sentences; this is however not self-evident.
A sentence may relate to a more remote sentence (i-5, for instance),
or to a block of sentences; see \cite{KuroNaga:discstruc} for a more
plausible model.
We found however that an online computer interface that would allow
the user to specify the target of a relation to this extent would become
prohibitively complicated.
The evaluator's task would involve so much juggling with relations and
attaining such a deep understanding of the text that it would in the
end have a negative effect on the reproducability and
evaluator-independence of the results.

\mbox{}\\{\bf Designing the dialog}\\
We believe that this is a trial-and-error process which will have to
be guided by the outcome of experiments; more about this will follow
below.

\section{The experiments}

We decided that experiments needed to establish three qualities of
this system.

\begin{description}
\item[Evaluator-independence]
Given a text in one language, different evaluators should produce the
same connectivity profile.
\item[Language-independence]
Given a ``perfectly'' translated text, its connectivity
profile should turn out the same as that of the original.
\item[Quantifiability]
Given translations of varying quality, the degree of correspondence in 
the connectivity profiles must be shown to correspond to the quality of 
the translation.
\end{description}

\noindent
But first we conducted a preliminary experiment.

\subsection{Experiments with the dialog}

Our first experiments (Japanese only) concerned the
conjunct-determining dialogs.
We implemented 3 interfaces, each comprising the same 61 conjuncts
spread over 9 categories:
one (A) based on categories (the subjects got a list of categories in
the first screen, and if they clicked one they got the conjuncts in
that category on the second screen);
one (B) based on the conjuncts themselves (the subjects just got the
whole list of conjuncts, spread over a couple of screens, without
elaboration);
and one (C) with questions (3 answers to choose from on the first
screen, one of these leads to a second question with 4 answers, all
other links lead to sets of conjuncts).

Subjects were assigned an interface, given a 9-sentence text and
asked to connect the sentences,
without however performing topic/comment extraction. 
A fourth group was asked to use interface C, but also to extract topic
and comment before connecting the sentences (D).
The results are given in table \ref{E1}.
The mean of the evaluators' choices was computed by transforming the
results into numbers (if 7 out of 10 evaluators chose category X, 
2 chose Y, and 1 Z, then this would result in the values \{1 1 1 1
1 1 1 2 2 3\}), and inputting these numbers into the following
formula.

\[ \mu^{e}_{2} = \frac{1}{n}\sum_{i=1}^{n}(x_{i} - \overline{x})^{2} \]

\begin{table}
\begin{center}
\begin{tabular}{|l||r|r|r|r|} \hline
& A & B & C & D \\ \hline \hline
mean & 0.43 & 0.46 & 0.37 & 0.32 \\ \hline
time (m:s) & 13:27 & 14:16 & 12:32 & 41:23 \\ \hline
backtracks & 11.8\hspace*{1.7mm} & 9.8\hspace*{1.7mm} &
4.4\hspace*{1.7mm} & 2.6\hspace*{1.7mm} \\ \hline
\hline
\end{tabular}
\caption{Results of the first experiment\label{E1}}
\end{center}
\end{table}

\noindent
We might add that subjects using interfaces A and B were more likely
to choose ``safe''
(ambiguous, vague)
conjuncts such as `soshite'
(and then), and also --- for what it's worth --- complained more.

\noindent
To be quite honest this experiment was too small in scale to allow
scientific conclusions (20 people participated),
but we went ahead anyway and concluded that
a) the project showed promise,
b) interface C was the way to go,
c) topic/comment extraction was important, but
d) it was also costly (took three times as long!) so we'd stick to the
`lazy' evaluation for further experiments.

\subsection{Validation experiments}

For the second set of experiments, we designed identical
interfaces for English and Japanese.
There was only one question, with 6 answers, and all of these led to
a screen with conjuncts to choose from, never more than 8 on a screen.
The set of conjuncts was designed to be minimal (no redundancies, no
ambiguous conjuncts); there were 32 of them, spread over 11 categories 
(cf.\ \S~\ref{Imp}).

An original English text was chosen (A);
then a ``perfect'' (but aligned) Japanese translation was produced (B);
and finally two
``less-than-perfect'' translations were contrived (C was raw MT
output, D was output from a tuned MT system the
understandability of which had been determined by independent
experiments to be halfway between B and C --- level 3 in
\cite{Fuji:genshoTK96}).
The sizes of the subject groups are given in table~\ref{E2} between
parentheses.
Distribution means were computed both for categories and for
conjuncts.

\begin{table}
\begin{center}
\begin{tabular}{|l||r|r|r|r|} \hline
& A (14) & B (13) & C (7) & D (7) \\ \hline \hline
mean(cat) & 0.52 & 0.60 & 0.89 & 0.69 \\ \hline
mean(con) & 1.99 & 2.66 & 2.20 & 1.76 \\ \hline
time (m:s) & 15:44 & 19:38 & 13:40 & 14:42 \\ \hline
backtracks & 4.6\hspace*{1.7mm} & 9.8\hspace*{1.7mm} &
6.7\hspace*{1.7mm} & 6.1\hspace*{1.7mm} \\ \hline \hline
\end{tabular}
{\flushright\footnotesize NB: (C+D) mean(cat) = 1.65, mean(con) = 4.91.\\}

\caption{Experiment results for the various texts\label{E2}}
\end{center}
\end{table}

\begin{table}
\begin{center}
\begin{tabular}{|l||r|r|r|r|} \hline
& A+B & A+C & A+D & A+C+D \\ \hline \hline
mean(cat) & 0.73 & 0.98 & 1.02 & 1.59 \\ \hline
mean(con) & 4.62 & 4.50 & 4.29 & 7.91 \\ \hline \hline
\end{tabular}
\caption{Combined experiment results\label{E22}}
\end{center}
\end{table}

\section{Discussion}

The category means basically follow expectations.
Those of C and D come out a bit low, but the combined mean for C+D
suggests that this may be partly due to the size of the sample.
The conjunct mean of B is very high; it is not clear why.
It must be noted that the evaluators were totally untrained; in the
context of the intended use of this method, requiring a certain level
of training seems acceptable and this would surely bring results
closer to the goal of evaluator independence.
However, we also observed several instances where the choice of a
conjunct was dictated by the evaluator's prior knowledge (or lack of
it) of the subject area; this is a discrepancy we cannot resolve.

The cross-linguistic category mean for A+B is significantly lower
than that of A+C and A+D.
The conjunct mean is rather high: this is probably due to the
unexplained high conjunct mean for B.
The conjunct means of A+C and A+D seem to correlate with the number of
unintelligible sentences in the machine-translated texts.
Again the means of A+C+D are fairly enormous, indicating that
size is still a factor.

A rather unsettling result, however, was that the most-chosen
sentence connector was identical across texts for almost each of the
sentence pairs.
This suggests that reducing evaluator dependence will lower {\em all}
means, which would defeat the purpose of this research.

In conclusion, we feel justified in hoping that the goals of
evaluator-independence and language-independence are reachable through
judicious tuning of the current system.
The project has also been successful in that it has yielded a wealth
of interesting data about sentence connections.
It is doubtful however that the approach will give a useful indication 
of translation quality.

\end{document}